\documentclass{article}
\usepackage[margin=0.85in]{geometry}
\usepackage[T1]{fontenc}
\usepackage{graphicx}
\usepackage{booktabs}
\usepackage{amsmath}
\usepackage{array}
\usepackage{microtype}
\usepackage[authoryear,round]{natbib}
\usepackage{placeins}
\usepackage{cuted}
\usepackage{capt-of}
\usepackage{float}
\usepackage{stfloats}
\usepackage[hidelinks,hypertexnames=false,plainpages=false]{hyperref}


\newcommand{\FullPageFigureGraphic}[1]{%
  \makebox[\textwidth][c]{%
    \includegraphics[width=0.98\paperwidth,height=0.86\textheight,keepaspectratio]{#1}%
  }%
}

\newcommand{\ActualPredictedHistogram}{%
  \IfFileExists{figures/actual_predicted_histograms.pdf}{%
    \FullPageFigureGraphic{figures/actual_predicted_histograms.pdf}%
  }{%
    \FullPageFigureGraphic{actual_predicted_histograms.pdf}%
  }%
}

\newlength{\CombinedAdmissionsPanelHeight}
\newcommand{\CombinedAdmissionsPanelGraphic}[1]{%
  \makebox[\textwidth][c]{%
    \includegraphics[width=1.05\textwidth,height=\CombinedAdmissionsPanelHeight,keepaspectratio]{#1}%
  }%
}

\newcommand{\DataSplittingTimelineFigure}{%
  \IfFileExists{figures/data_splitting_timeline_panel.pdf}{%
    \CombinedAdmissionsPanelGraphic{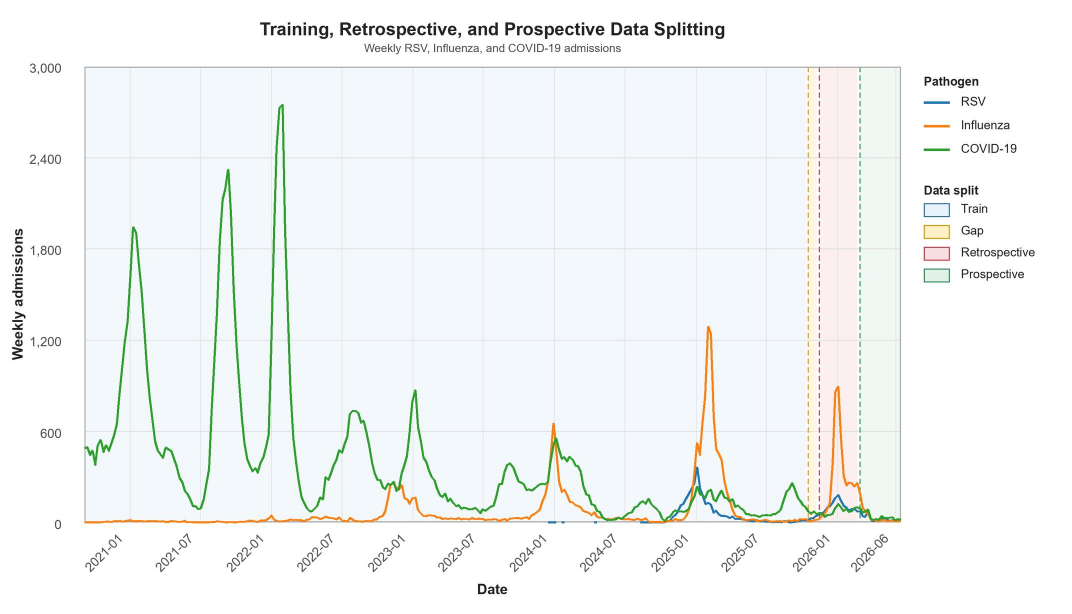}%
  }{%
    \IfFileExists{data_splitting_timeline_panel.pdf}{%
      \CombinedAdmissionsPanelGraphic{data_splitting_timeline_panel.pdf}%
    }{%
      \fbox{\parbox{0.9\textwidth}{\centering Data-splitting timeline figure placeholder.}}%
    }%
  }%
}

\newcommand{\CombinedAdmissionsPaMaeFigure}{%
  \IfFileExists{figures/combined_admissions_pa_mae_no_yaxis_wide_bold_xticks.pdf}{%
    \CombinedAdmissionsPanelGraphic{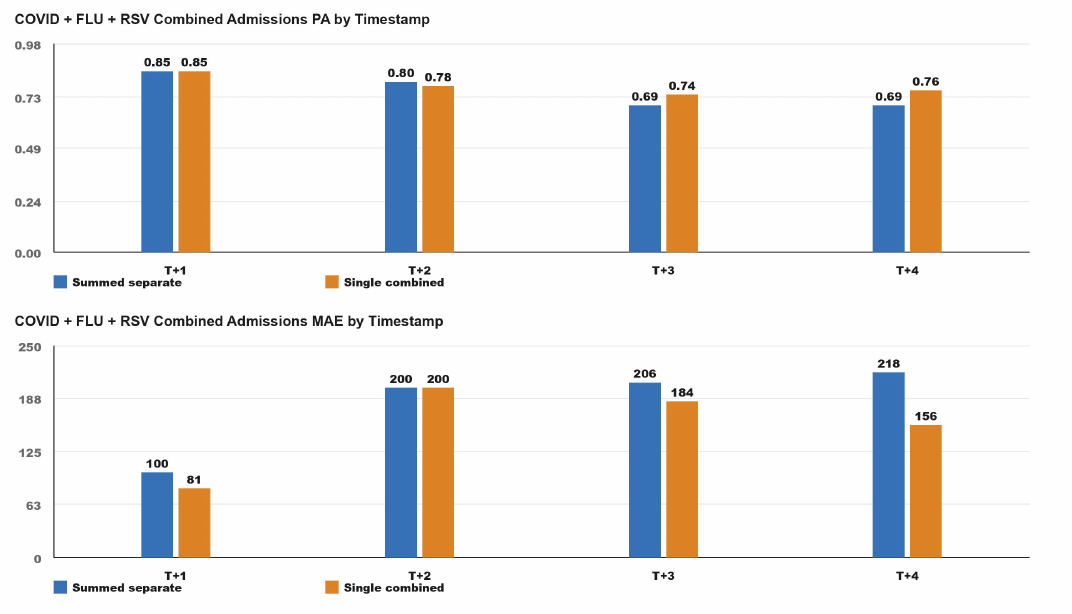}%
  }{%
    \IfFileExists{combined_admissions_pa_mae_no_yaxis_wide_bold_xticks.pdf}{%
      \CombinedAdmissionsPanelGraphic{combined_admissions_pa_mae_no_yaxis_wide_bold_xticks.pdf}%
    }{%
      \IfFileExists{figures/combined_admissions_pa_mae_no_yaxis_wide_bold.pdf}{%
        \CombinedAdmissionsPanelGraphic{figures/combined_admissions_pa_mae_no_yaxis_wide_bold.pdf}%
      }{%
        \IfFileExists{combined_admissions_pa_mae_no_yaxis_wide_bold.pdf}{%
          \CombinedAdmissionsPanelGraphic{combined_admissions_pa_mae_no_yaxis_wide_bold.pdf}%
        }{%
          \IfFileExists{figures/combined_admissions_pa_mae_no_yaxis_wide.pdf}{%
            \CombinedAdmissionsPanelGraphic{figures/combined_admissions_pa_mae_no_yaxis_wide.pdf}%
          }{%
            \IfFileExists{combined_admissions_pa_mae_no_yaxis_wide.pdf}{%
              \CombinedAdmissionsPanelGraphic{combined_admissions_pa_mae_no_yaxis_wide.pdf}%
            }{%
              \IfFileExists{figures/combined_admissions_pa_mae_no_yaxis.pdf}{%
                \CombinedAdmissionsPanelGraphic{figures/combined_admissions_pa_mae_no_yaxis.pdf}%
              }{%
                \IfFileExists{combined_admissions_pa_mae_no_yaxis.pdf}{%
                  \CombinedAdmissionsPanelGraphic{combined_admissions_pa_mae_no_yaxis.pdf}%
                }{%
                  \IfFileExists{figures/combined_admissions_pa_mae_no_yaxis.png}{%
                    \CombinedAdmissionsPanelGraphic{figures/combined_admissions_pa_mae_no_yaxis.png}%
                  }{%
                    \fbox{\parbox{0.75\textwidth}{\centering Placeholder for combined admissions PA and MAE figure.}}%
                  }%
                }%
              }%
            }%
          }%
        }%
      }%
    }%
  }%
}

\newcommand{\CombinedAdmissionsProspectivePaMaeFigure}{%
  \IfFileExists{figures/combined_admissions_prospective_pa_mae_no_yaxis_wide_bold_xticks.pdf}{%
    \CombinedAdmissionsPanelGraphic{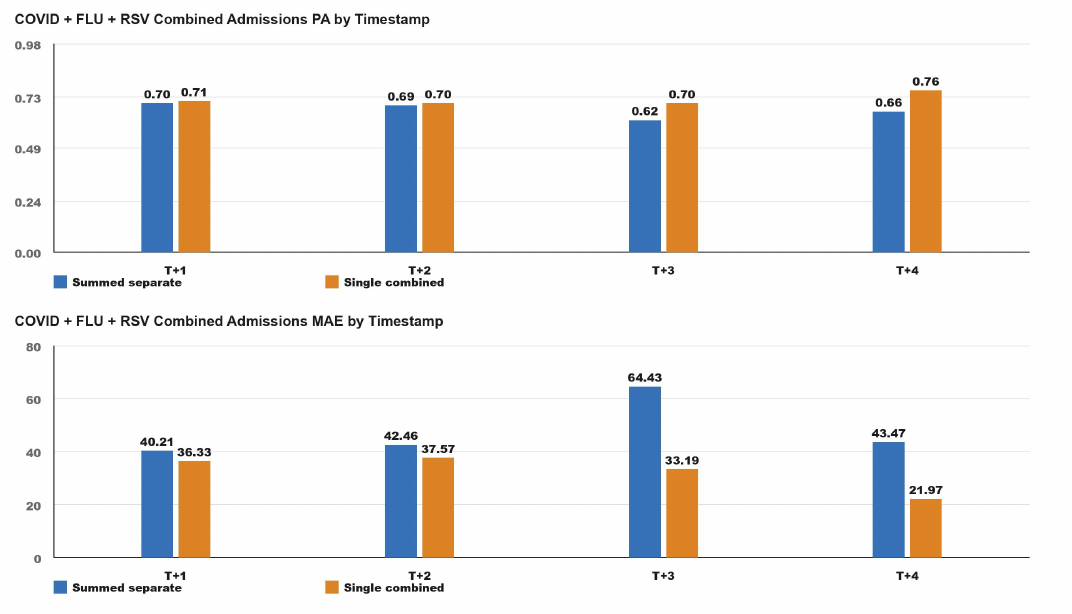}%
  }{%
    \IfFileExists{combined_admissions_prospective_pa_mae_no_yaxis_wide_bold_xticks.pdf}{%
      \CombinedAdmissionsPanelGraphic{combined_admissions_prospective_pa_mae_no_yaxis_wide_bold_xticks.pdf}%
    }{%
      \IfFileExists{figures/combined_admissions_prospective_pa_mae_no_yaxis_wide.pdf}{%
        \CombinedAdmissionsPanelGraphic{figures/combined_admissions_prospective_pa_mae_no_yaxis_wide.pdf}%
      }{%
        \IfFileExists{combined_admissions_prospective_pa_mae_no_yaxis_wide.pdf}{%
          \CombinedAdmissionsPanelGraphic{combined_admissions_prospective_pa_mae_no_yaxis_wide.pdf}%
        }{%
          \IfFileExists{figures/combined_admissions_prospective_pa_mae_no_yaxis.pdf}{%
            \CombinedAdmissionsPanelGraphic{figures/combined_admissions_prospective_pa_mae_no_yaxis.pdf}%
          }{%
            \IfFileExists{combined_admissions_prospective_pa_mae_no_yaxis.pdf}{%
              \CombinedAdmissionsPanelGraphic{combined_admissions_prospective_pa_mae_no_yaxis.pdf}%
            }{%
              \IfFileExists{figures/combined_admissions_mae_no_yaxis.pdf}{%
                \CombinedAdmissionsPanelGraphic{figures/combined_admissions_mae_no_yaxis.pdf}%
              }{%
                \IfFileExists{combined_admissions_mae_no_yaxis.pdf}{%
                  \CombinedAdmissionsPanelGraphic{combined_admissions_mae_no_yaxis.pdf}%
                }{%
                  \IfFileExists{figures/combined_admissions_mae_no_yaxis.png}{%
                    \CombinedAdmissionsPanelGraphic{figures/combined_admissions_mae_no_yaxis.png}%
                  }{%
                    \fbox{\parbox[c][\CombinedAdmissionsPanelHeight][c]{0.75\textwidth}{\centering Placeholder for prospective PA and MAE figure. Upload \texttt{combined\_admissions\_prospective\_pa\_mae\_no\_yaxis\_wide\_bold\_xticks.pdf}.}}%
                  }%
                }%
              }%
            }%
          }%
        }%
      }%
    }%
  }%
}

\newcommand{\OperationalForecastBarChart}{%
  \IfFileExists{figures/actual_predicted_histograms.pdf}{%
    \FullPageFigureGraphic{figures/actual_predicted_histograms.pdf}%
  }{%
    \FullPageFigureGraphic{actual_predicted_histograms.pdf}%
  }%
}

\newcommand{\PathogenCompositionGraphic}{%
  \includegraphics[width=0.95\textwidth]{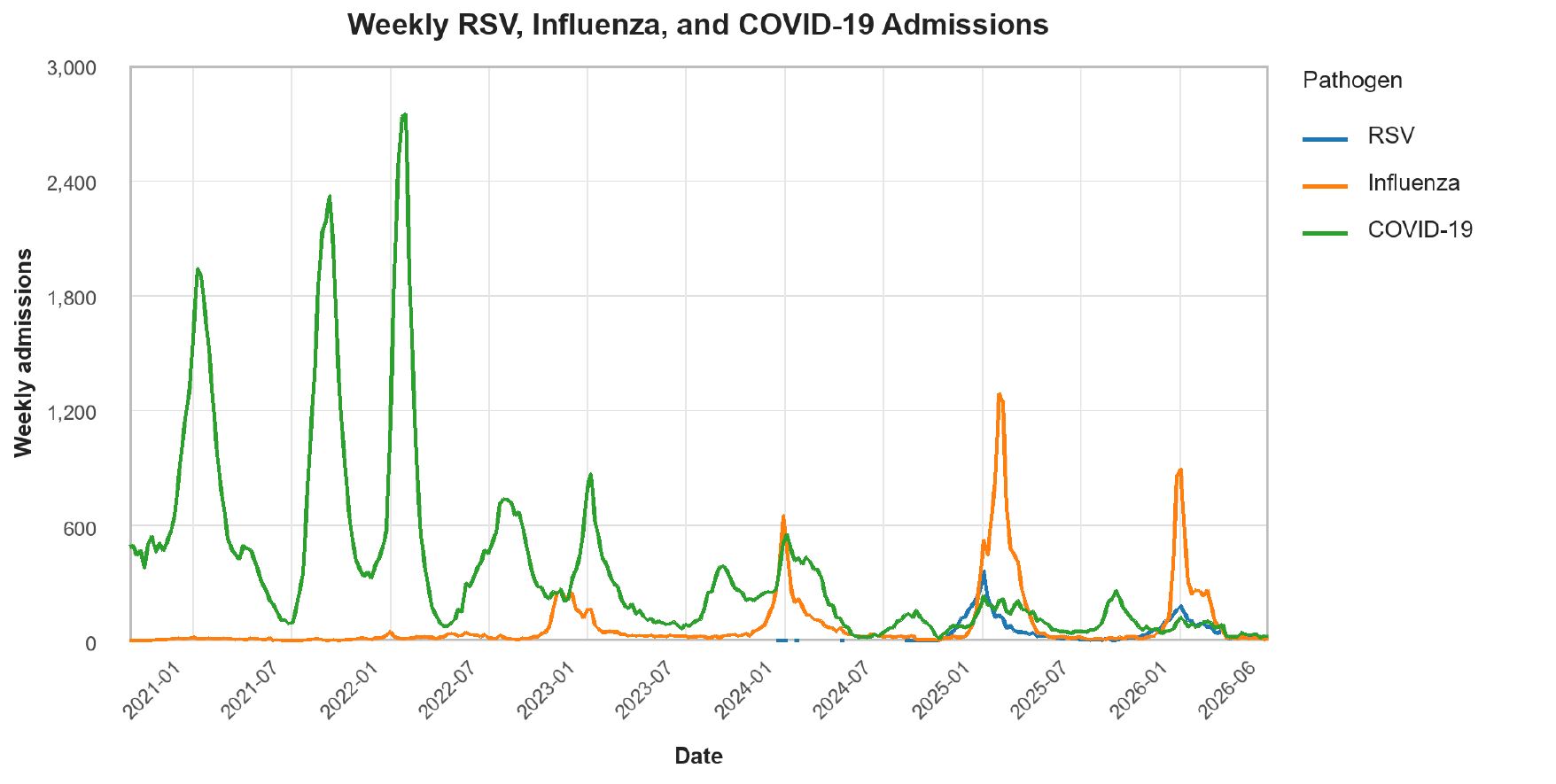}%
}

\title{Beyond Single-Pathogen Forecasts: Predicting Shared Respiratory Hospital Burden from Influenza, COVID-19, and RSV}
\author{%
\parbox{0.92\textwidth}{\centering
Gaurob Saha, Minjae Woo, Lior Rennert\\[0.35em]
Department of Public Health Sciences and Center for Public Health Modeling and Response,\\
Clemson University, Clemson, SC 29601, USA\\[0.35em]
\small E-mail addresses: gsaha@clemson.edu, mwoo@clemson.edu, liorr@clemson.edu}}
\date{}

\begin{document}

\maketitle

\begin{abstract}
Co-circulating influenza, COVID-19, and respiratory syncytial virus (RSV) create a shared hospital-admission
burden, because beds, staff, testing capacity, and respiratory support are allocated to total demand rather than to
one pathogen at a time. This makes respiratory forecasting a target-construction problem as well as a modeling
problem: disease-specific forecasts preserve attribution, whereas hospitals also need a single planning signal for
combined burden. We developed a supervised 1- to 4-week-ahead framework for weekly admissions and compared two
total-burden strategies: summing independently generated pathogen-specific forecasts and forecasting the combined
admission target directly. Evaluation included retrospective held-out testing and prospective operational forecasts
issued before the corresponding admission weeks were observed, with performance summarized using MAE and Percent
Agreement (PA; the smaller of model-generated and observed counts divided by the larger). In retrospective
total-burden evaluation, the combined-burden model had PA values of 0.85, 0.78, 0.74, and 0.76 and MAE values of 80.63, 200.09,
183.97, and 156.13 from week 1 to week 4 ahead, compared with summed-forecast PA values of 0.85, 0.80, 0.69, and
0.69 and MAE values of 100.14, 200.06, 205.65, and 217.57. Prospectively, the combined-burden model had higher PA and
lower MAE at all four horizons, with the largest separations at weeks 3 and 4 ahead. These results support layered
respiratory forecasting: pathogen-specific forecasts remain necessary for etiologic interpretation, while
combined-burden forecasts better align with hospital preparedness and shared-capacity planning.
\end{abstract}

\noindent\textbf{Keywords:} Respiratory hospital burden; Influenza; COVID-19; RSV; Multi-pathogen forecasting; Hospital preparedness.

\clearpage
\twocolumn
\raggedbottom

\section{Introduction}

Respiratory viruses are a major cause of seasonal hospital burden, with influenza, coronavirus disease 2019
(COVID-19), and respiratory syncytial virus (RSV) contributing substantially to emergency department visits,
inpatient admissions, intensive care utilization, and mortality
\citep{Bi2025,Dewey2025,Mellor2025,Yang2024RSV,Gebremedhin2022RSV}. During the
2023--2024 respiratory season in the United States, these three pathogens accounted for more than one million
hospitalizations, with combined daily admissions exceeding 10,000 during the seasonal peak \citep{Bi2025}. Such
surges place considerable pressure on healthcare systems by increasing demand for inpatient beds, intensive
care units, ventilators, diagnostic testing, healthcare personnel, and other hospital resources
\citep{Mellor2025,Osthus2022,Gebremedhin2022RSV}. Because influenza, COVID-19, and RSV frequently circulate during the same respiratory season, they compete for these shared resources,
requiring hospitals to manage the overall respiratory disease burden rather than each pathogen independently
\citep{Bi2025,Dewey2025,Mellor2025}.

Numerous forecasting models have been developed to estimate short-term hospital admissions for influenza,
COVID-19, and, more recently, RSV
\citep{Ray2025Flusion,Klein2023,Mellor2025,Yang2024RSV}. These studies have demonstrated the
utility of statistical models, machine learning methods, ensemble approaches, and multi-source surveillance
data for supporting near-term public health planning and hospital preparedness
\citep{Osthus2022,Ray2025Flusion,Park2024DCCFeatureSelection,Yoon2026ILIEnsemble,Mellor2025}. Despite these advances, respiratory
disease forecasting continues to be largely pathogen-specific. Influenza, COVID-19, and RSV are typically
modeled as separate prediction problems, each developed using its own surveillance data, preprocessing
strategy, feature engineering procedure, model optimization, calibration, and evaluation framework
\citep{Osthus2022,Ray2025Flusion,Klein2023,Yang2024RSV}. While this approach may provide useful
forecasts for individual diseases, it implicitly treats respiratory pathogens as independent processes, even
though they coexist within the same healthcare system. It also does not align with the way respiratory-season
burden is observed or managed in hospitals, where multiple pathogens collectively determine demand for the same
beds, staff, diagnostic capacity, and respiratory support \citep{Bi2025,Dewey2025,Mellor2025}.

\begin{figure*}[!t]
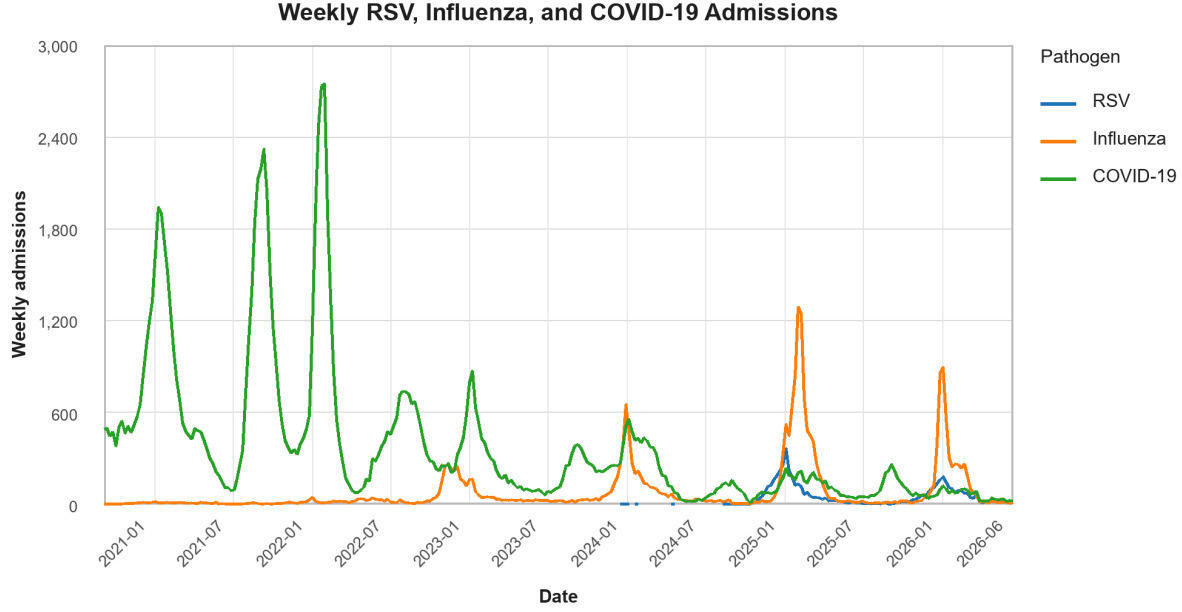

\centering
\PathogenCompositionGraphic
\caption{Weekly respiratory hospital admissions by pathogen across respiratory seasons. The figure shows the
changing relative contributions of COVID-19, influenza, and RSV to the overall respiratory hospitalization
burden, motivating a unified multi-pathogen forecasting framework.}
\label{fig:pathogen_composition}
\end{figure*}

This pathogen-specific forecasting paradigm has important limitations. Influenza, COVID-19, and RSV do not
circulate in isolation, and their observed hospital burden may be shaped by overlapping symptoms, testing
behavior, healthcare-seeking patterns, population immunity, vaccination patterns, and concurrent respiratory
virus activity \citep{Bi2025,Dewey2025,Mellor2025,Gebremedhin2022RSV,Yang2024RSV}. In our surveillance
data, the relative contribution of each pathogen changed substantially across respiratory seasons, with COVID-19
dominating earlier periods, influenza becoming the largest contributor during selected winter peaks, and RSV
adding a distinct seasonal component to the overall hospitalization burden
(Fig.~\ref{fig:pathogen_composition}). This changing composition illustrates why a model trained only on one
pathogen remains unaware of the contemporaneous respiratory disease environment in which that pathogen occurs.
Consequently, simply adding predictions from independently developed pathogen-specific models is not equivalent
to forecasting total respiratory burden; the summed estimate inherits separate model assumptions, separate
calibration behavior, and separate error structures while ignoring information contained in the collective
activity of respiratory pathogens. This creates an urgent gap in understanding how discrepancies between
pathogen-specific forecasts and multi-pathogen respiratory burden forecasts may lead hospitals to overestimate
or underestimate the resources needed for near-term respiratory-season planning \citep{Mellor2025,Murph2026EpiFFORMA}.

In this paper, we describe a unified multi-pathogen forecasting framework for short-term respiratory hospital
admissions, with the purpose of evaluating whether respiratory burden is better estimated using a single
forecasting framework than by aggregating independently developed pathogen-specific models. The presented
analysis compares these two operational forecasting paradigms through retrospective evaluation and prospective
deployment across one- to four-week forecasting horizons. Beyond comparing forecasting accuracy, this study
examines whether a unified forecasting strategy provides a more representative estimate of respiratory disease
burden while enabling a single operational workflow for influenza, COVID-19, and RSV. The proposed framework
has the potential to support hospital preparedness by improving estimation of overall respiratory demand,
simplifying operational forecasting, and providing a scalable foundation for future multi-pathogen surveillance
and response \citep{Osthus2022,Mellor2025,Murph2026EpiFFORMA}.

\section{Methodology}

The methodological workflow was organized as a sequential forecasting pipeline. Weekly surveillance and
hospital-admission data were first harmonized across sources, then used to construct disease-specific and
combined respiratory-burden targets. For each target and horizon, the pipeline applied horizon shifting, feature
selection, feature engineering, model tuning, adaptive forecast adjustment, and benchmark comparison before
retrospective and prospective scoring.

\subsection{Data Sources}

Weekly respiratory surveillance and hospital admission data were assembled separately for influenza,
COVID-19, and RSV over a common experimental period from 2020-09-05 through 2026-06-13. These weekly aggregate
data supported two related forecasting targets: disease-specific admissions and the aligned combined respiratory
admission burden used for hospital-preparedness planning.

Three source types were used. CDC NHSN served as the primary source for weekly admission targets and broader
statewide hospital-burden indicators. MUSC and PRISMA supplied local health-system surveillance predictors. South
Carolina Revenue and Fiscal Affairs (RFA) RSV hospitalization data were used only to bridge the historical RSV
target before CDC RSV admissions were consistently available. This role-based harmonization was needed because no
single stream provided complete disease-specific admission targets and local surveillance signals across the full
study period.

For influenza and COVID-19, the modeling target was the CDC NHSN weekly total admission count for the
corresponding disease. For RSV, CDC NHSN weekly total admission count was used once the CDC RSV target was available for the modeling period. Before that point, the CDC RSV target was missing or not consistently usable for
training, so the RFA weekly hospitalization series was
used to extend the historical RSV target. Within the study window, RFA weekly hospitalizations contributed
historical RSV target weeks from 2020-09-05 through 2024-09-21 after conversion to the CDC admission scale.
CDC Total RSV Admissions was used for RSV target weeks from 2024-09-28 onward, including the test weeks.

The MUSC and PRISMA files provided the same local weekly indicator types for each disease: disease-related
diagnoses, emergency-department visits, tests performed, positive tests, vaccines recorded during the reporting
week, season-to-date vaccinations, and total disease-related clinical encounters. Additional CDC predictors
provided broader hospital-burden context when selected by the pipeline, including admission rates per 100,000
population, pediatric admission and hospitalization counts, ICU hospitalization counts, bed-occupancy
percentages, and cumulative seasonal admission counts. This source mix is motivated by prior work showing that
local health-system and other near-real-time surveillance streams can provide actionable signals for infectious
disease burden when broader public health datasets are delayed or too coarse for local planning
\citep{Ahammed2025LocalHealthSystem,Tabassum2026Wastewater,Klein2023}.

\subsection{Data Preprocessing}

The preprocessing workflow harmonizes the source files before modeling. The disease-specific data sources are
merged on week:

\begin{equation}
    D =
    \left(
    D_{\mathrm{MUSC}}
    \bowtie_{\mathrm{Week}}
    D_{\mathrm{PRISMA}}
    \right)
    \bowtie_{\mathrm{Week}}
    D_{\mathrm{CDC}}.
\end{equation}

For RSV, the MUSC, PRISMA, and CDC files remained the feature sources in the merged modeling dataset. The RFA
file was linked by Week only to construct the historical RSV target before the CDC RSV target period, after
converting RFA weekly hospitalizations to the CDC admission scale.

The RFA-to-CDC conversion was developed using only the weeks when both CDC weekly RSV admissions and RFA
weekly RSV hospitalizations were available. This shared RFA--CDC overlap spanned 2024-09-28 through
2025-02-15 and provided 21 paired weekly observations.

Four candidate conversion rules were evaluated in this overlap period. Each rule converted RFA weekly RSV
hospitalizations into a CDC-scale RSV admission value. Simple ratio scaling used a single RFA-to-CDC
multiplier; threshold-ratio scaling first mapped low RFA values to 0 and then applied a multiplier to the
remaining values; RFA-only linear regression estimated CDC admissions from RFA hospitalizations alone; and the
RFA-plus-indicators regression added selected PRISMA RSV inpatient-hospitalization, emergency-department,
and positive-test indicators to the RFA value. Because the overlap period was short, the comparison used
leave-one-out cross-validation across the 21 overlap weeks. In each fold, one overlap week was withheld, each
candidate conversion rule was estimated from the other 20 overlap weeks, and the withheld RFA value was
converted to the CDC scale and compared with the observed CDC RSV admission value for that same week.

The bridge methods were compared using Percent Agreement (PA), because the conversion step was intended to
preserve proportional agreement between the reconstructed RSV target and the CDC admission scale rather than
optimize absolute error alone. In general, PA measures relative closeness by comparing the smaller value in a
predicted-observed pair with the larger value, so values closer to 1 indicate stronger agreement. For each
withheld overlap week, PA was calculated as the smaller of the observed CDC value and converted RFA value
divided by the larger value. The
leave-one-out PA values were 0.630 for simple ratio scaling, 0.761 for threshold-ratio scaling, 0.633 for
RFA-only regression, and 0.697 for RFA plus selected indicators. Based on this criterion, the threshold-ratio
bridge was selected because it had the highest PA and provided a transparent single-series conversion rule for
historical RFA admissions. The selected bridge was fixed before retrospective and prospective evaluation.

After selection, the final threshold and scaling values were calculated from the full overlap period. The overlap
pattern showed that a simple multiplication rule would not be adequate for low RFA values. During the first
three overlap weeks, CDC RSV admissions remained 0 while RFA reported 65, 69, and 74 weekly hospitalizations.
The first positive CDC RSV admission value occurred when RFA reached 103. The zero threshold was therefore set
at the midpoint between the largest RFA value paired with a CDC value of 0 and the smallest RFA value paired
with a positive CDC value: (74 + 103)/2 = 88.5.

\[
\tilde{y}_t^{(\mathrm{RSV})}
=
\begin{cases}
0, & r_t < 88.5, \\
0.1969157929\,r_t, & r_t \ge 88.5,
\end{cases}
\]

where $r_t$ is the RFA weekly RSV hospitalization count and $\tilde{y}_t^{(\mathrm{RSV})}$ is the converted
CDC-scale RSV admission target. The scaling ratio was estimated from the 18 overlap weeks in which both CDC
and RFA values were positive. For each positive overlap week, the weekly ratio was calculated as CDC weekly RSV
admissions divided by RFA weekly RSV hospitalizations. The ratios were then sorted, the single lowest and
single highest ratios were removed, and the remaining 16 ratios were averaged. This 10\% trimmed mean was used
to reduce sensitivity to unusually high or low overlap weeks. The converted RFA target was used only before
2024-09-28; CDC Total RSV Admissions was used once available and for all retrospective and prospective scoring.

After the disease-specific sources were merged and the active disease target was constructed, including the
RFA-to-CDC conversion for RSV, the remaining preprocessing steps were applied uniformly across diseases. All
non-date fields were converted to numeric values after standardizing common missing-value tokens, including
empty strings, null, and dash placeholders. The merged dataset was then sorted by Week. Rows without an observed target value for the active disease were removed from the modeling frame.

\subsection{Forecast Target Construction and Burden Paradigms}

This study treats target construction as part of the forecasting question. Single-pathogen targets preserve
etiologic attribution for influenza, COVID-19, and RSV. Summed forecasts estimate total burden only after
separate pathogen-specific predictions have been generated. The combined-burden paradigm instead defines total
respiratory admissions as the hospital planning target before model fitting and evaluation. This distinction is
central because disease-specific interpretation and shared-capacity planning answer related but different
operational questions. In retrospective evaluation, both paradigms are scored on out-of-sample test weeks for
which future admission targets are already observed. In prospective forecasting, the same target definitions are
used to generate forecasts from the most recent available predictor rows before the future admission counts are
observed and scored.

\subsubsection{Single-Pathogen Target Construction}

For the single-pathogen paradigm, each disease was modeled as a separate weekly admission burden. These targets
support pathogen attribution and allow influenza, COVID-19, and RSV forecasts to be evaluated individually. Let
$y_t^{(d)}$ denote the observed admission count at week $t$ for disease
$d \in \{\mathrm{Influenza}, \mathrm{COVID\mbox{-}19}, \mathrm{RSV}\}$. The disease-specific targets were

\begin{equation}
    y_t^{(\mathrm{Influenza})}, \qquad
    y_t^{(\mathrm{COVID\mbox{-}19})}, \qquad
    y_t^{(\mathrm{RSV})}.
\end{equation}

For influenza and COVID-19, these targets were the corresponding CDC NHSN weekly admission counts. For RSV,
the target combined the converted RFA-scale historical values before the CDC RSV target period with CDC NHSN
RSV admissions once the CDC target was available. Each disease-specific model used its own target, selected
predictors, engineered features, and horizon-specific training data. The resulting disease-specific forecasts
can be evaluated separately or summed after prediction to obtain an aggregate estimate from independently
modeled pathogens:

\begin{equation}
    \hat{y}_{t,h}^{(\mathrm{sum})}
    =
    \hat{y}_{t,h}^{(\mathrm{Influenza})}
    +
    \hat{y}_{t,h}^{(\mathrm{COVID\mbox{-}19})}
    +
    \hat{y}_{t,h}^{(\mathrm{RSV})}.
\end{equation}

\subsubsection{Multi-Pathogen Target Construction}

For the multi-pathogen paradigm, the weekly respiratory burden target was constructed by aligning the
disease-specific admission targets by week and summing them:

\begin{equation}
\begin{aligned}
    y_t^{(\mathrm{Combined})}
    &=
    y_t^{(\mathrm{Influenza})}
    +
    y_t^{(\mathrm{COVID\mbox{-}19})} \\
    &\quad+
    y_t^{(\mathrm{RSV})}.
\end{aligned}
\end{equation}

This target represents the total weekly hospital admission burden attributable to the three respiratory
pathogens in the study. In contrast to the summed single-pathogen forecast, the multi-pathogen target
construction defines the shared hospital-burden signal before horizon shifting, model fitting, and evaluation. 

\subsection{Performance Evaluation}
\label{sec:performance_evaluation}

Performance was evaluated using point-prediction metrics applied consistently across retrospective estimation
and prospective forecasting. Because the target variables are weekly admission counts, all model outputs were
clipped to be nonnegative before scoring:

\begin{equation}
    \tilde{y}_i
    =
    \max\{\hat{y}_i, 0\},
\end{equation}

where $\hat{y}_i$ is the raw predicted value and $\tilde{y}_i$ is the scored prediction. For an evaluated set
$\mathcal{S}_{q,h}$ of observations for target $q$ and horizon $h$, mean absolute error (MAE) was defined as

\begin{equation}
    \mathrm{MAE}_{q,h}
    =
    \frac{1}{|\mathcal{S}_{q,h}|}
    \sum_{i \in \mathcal{S}_{q,h}}
    |y_i - \tilde{y}_i|.
\end{equation}

MAE is measured in admissions, so lower values indicate smaller absolute prediction errors. Percent Agreement
(PA) was used as a scale-free measure of proportional closeness between observed and predicted admissions:

\begin{equation}
    \mathrm{PA}_{q,h}
    =
    \frac{1}{|\mathcal{S}_{q,h}|}
    \sum_{i \in \mathcal{S}_{q,h}}
    A_i,
    \label{eq:percent-agreement}
\end{equation}

where
\begin{equation}
    A_i
    =
    \begin{cases}
    1, & y_i = 0 \ \mathrm{and} \ \tilde{y}_i = 0, \\
    \dfrac{\min(y_i,\tilde{y}_i)}
    {\max(y_i,\tilde{y}_i)}, & \mathrm{otherwise}.
    \end{cases}
\end{equation}

PA ranges from 0 to 1, with 1 indicating exact agreement and higher values indicating stronger proportional
agreement. Thus, PA captures proportional agreement, whereas MAE captures admission-count error directly
relevant to staffing and capacity planning. Metrics were computed separately by target and horizon. For
total-burden evaluation, summed
single-pathogen predictions were first added across influenza, COVID-19, and RSV and then compared with the
corresponding summed observed admissions; the combined-burden model was scored against the observed combined
respiratory-burden target. All performance metrics were calculated from point predictions; uncertainty intervals
or full predictive distributions were not evaluated in this study.

\subsection{Horizon-Specific Modeling Workflow}

For near-term respiratory-burden planning, each admission target was converted into horizon-specific prediction
tasks for influenza, COVID-19, RSV, and the combined respiratory-burden target across four forecast horizons.
Here, $T$ denotes the forecast origin; $T+1$ through $T+4$ denote forecasts 1 through 4 weeks ahead,
respectively.

The horizon-specific shifted-label construction was applied separately within each paradigm. In the
single-pathogen paradigm, separate shifted targets were created for influenza, COVID-19, and RSV. In the
multi-pathogen paradigm, the aligned total respiratory burden target was shifted directly. Let $y_t^{(q)}$
denote the observed weekly hospital burden value for target $q$, where
$q \in \{\mathrm{Influenza}, \mathrm{COVID\mbox{-}19}, \mathrm{RSV}, \mathrm{Combined}\}$. The label for target
$q$ and horizon $h$ is

\begin{equation}
    y_{t,h}^{(q)} = y_{t+h}^{(q)}.
\end{equation}

In implementation, this corresponds to shifting each target column upward by $h$ weeks so that predictors at
week $t$ are paired with admissions at week $t+h$.

Here, Target represents either an active disease target:
Total Influenza Admissions,
Total COVID-19 Admissions, or
Total RSV Admissions, or the aligned total respiratory burden target used in the multi-pathogen
paradigm. This construction
aligns each row of predictors at week $t$ with the future burden value at week $t+h$. Using one shifted-label
rule across all targets keeps disease-specific and combined-burden results comparable at each
horizon. It also preserves the forecast-time information set by pairing predictors from week $t$ or earlier with
future admissions. The shifted-label rule is handled
differently for retrospective testing and prospective forecasting. During retrospective model training and
testing, a row can be used only if its future target has already been observed. Therefore, after shifting the
target, the T+1 dataset excludes the final one row without an observed week-$t+1$ admission value, the T+2
dataset excludes the final two rows without observed week-$t+2$ values, and the same logic applies to T+3 and
T+4. These unlabeled end-of-series rows were excluded from retrospective training and scoring; scoring used
only labeled rows in the chronological test segment after the fixed temporal gap. For prospective forecasting, the
most recent row with a complete predictor vector was retained as the forecast-origin row, and
the trained horizon-specific models were applied to this row to generate forecasts before the target admission
weeks were observed. After the corresponding admission counts became available, the prospective forecasts were
scored using the same evaluation metrics.

\subsubsection{Feature Selection}

Feature selection is horizon-specific because the association between predictors and future admissions may
change as the forecast horizon increases. For each forecasting target and horizon, the data with shifted targets are
partitioned chronologically into a training segment, a fixed temporal gap, and a test segment:

\begin{equation}
    D_h =
    D_{h,\mathrm{train}}
    \;|\;
    D_{\mathrm{gap}}
    \;|\;
    D_{h,\mathrm{test}}.
\end{equation}

The fixed temporal gap corresponds to the skipped calendar weeks between the training and retrospective test
segments, not to the forecast horizon $h$. These rows are excluded from both model training and retrospective
testing to avoid temporal leakage. Predictors with any missing values in the training segment are removed based
on the training set only, not the gap or test periods. Candidate features are then screened using mutual information, followed by
variance inflation factor filtering.

\paragraph*{Mutual information (MI) score.} Mutual information was used to rank candidate predictors according to their nonlinear dependence with the
horizon-specific target. For candidate predictor $X_j$ and target $Y_h$, mutual information is defined as

\begin{equation}
    I(X_j;Y_h)
    =
    \int \int p(x_j,y_h)
    \log
    \left(
    \frac{p(x_j,y_h)}{p(x_j)p(y_h)}
    \right)
    dx_jdy_h.
\end{equation}

In the implementation, MI is estimated with mutual info regression using only the training data.
The top 40 predictors are retained for collinearity screening. In addition, the pipeline always appends these local hospital indicators to the selected feature list:
Weekly Inpatient Hospitalizations MUSC and
Weekly Positive Tests PRISMA. These forced-keep predictors, together with
the top 10 MI-ranked predictors, are protected during VIF filtering. This design
preserves clinically meaningful local signals even when feature rankings vary across small weekly training
samples. More broadly, local and multi-source infectious disease forecasting studies emphasize the value of
combining data-driven modeling with clinically relevant surveillance signals
\citep{Ahammed2025LocalHealthSystem,Wu2025Transformer,Hossain2025RtCases}.

\paragraph*{Variance inflation factor (VIF).} Following MI screening, variance inflation factor filtering is used to reduce multicollinearity among retained
features. For predictor $X_j$, let $R_j^2$ be the coefficient of determination obtained by regressing $X_j$ on
the remaining selected predictors. The VIF is

\begin{equation}
    \mathrm{VIF}_j = \frac{1}{1 - R_j^2}.
\end{equation}

High VIF values indicate that a feature is strongly explained by other selected predictors. The pipeline uses
a VIF threshold of 100. Features exceeding this threshold are removed iteratively, except for the
protected top 10 MI-ranked predictors and the forced-keep predictors. When multiple
removable features exceed the threshold, the implementation removes the feature with the lowest MI score
first; if multiple removable features have similarly low MI scores, the higher-VIF feature is removed first.
Although conventional VIF screening often uses lower cutoffs such as 5 or 10, these cutoffs are heuristic
rather than universal \citep{OBrien2007VIF}. Preliminary sensitivity checks using
thresholds of 5, 10, 15, 20, 25, and 50 retained substantially fewer predictors and reduced validation PA.
The threshold of 100 was therefore retained to remove only extreme redundancy before tree-ensemble prediction,
where collinearity mainly affects redundant split choices rather than coefficient interpretation.
This horizon-specific MI and VIF process means that different numbers of predictors can be selected for T+1,
T+2, T+3, and T+4, as shown in
Table~\ref*{tab:training_summary} for the disease-specific models and
Table~\ref*{tab:combined_training_summary} for the combined-burden model.
The resulting feature set is then passed to feature engineering.

\subsubsection{Feature Engineering}

As shown in Figure~\ref{fig:pathogen_composition}, weekly admissions exhibited respiratory-burden dynamics,
including recurrent seasonal peaks, pathogen-specific timing, and short-term persistence. These patterns
motivated lagged, momentum, smoothing, and seasonal predictors designed to capture recent burden, trend,
smoothed trajectory, and annual timing. This framing treats feature engineering as a way to represent
respiratory-burden dynamics in the surveillance series, rather than as a purely algorithmic expansion of
predictors. Lagged values and rolling means represent recent burden and smoothed trajectory, percentage
changes, ratios, and first- and second-order differences represent local trend and short-term momentum, and
Fourier terms represent annual seasonality.
This step is applied
after MI and VIF screening, so the feature expansion is performed only on the target- and horizon-specific
predictors retained for that model. Consequently, the number of engineered features can differ across T+1,
T+2, T+3, and T+4, as reported in Table~\ref*{tab:training_summary}. The
retained surveillance and hospital variables are treated as exogenous predictors. The observed admission
history is kept separately so that past admission values can be used, while the shifted future admission label
is kept only as the outcome to be predicted. This target label is never engineered as a predictor. These feature
families mirror common signals used in respiratory forecasting, where recent observed burden, lagged
surveillance indicators, and seasonal terms often explain different parts of the trajectory
\citep{Osthus2022,Ray2025Flusion,Morris2023WebSearchUncertainty}.

For each retained exogenous predictor $x_t$ observed at week $t$, the implementation first creates three
backward-looking lagged values:

\begin{equation}
    x_{t-k}, \qquad k = 1,2,3.
\end{equation}

These lagged values allow the model to use the recent history of diagnoses, visits, tests, positive tests,
hospital indicators, and CDC surveillance variables rather than the current week alone. Short-term relative
changes are encoded with safe percentage-change features:

\begin{align}
    \Delta^{\%}_{1}(x_t)
    &=
    \frac{x_t - x_{t-1}}{x_{t-1}}, \\
    \Delta^{\%}_{2}(x_t)
    &=
    \frac{x_{t-1} - x_{t-2}}{x_{t-2}}.
\end{align}

In the implementation, these percentage-change features are protected against zero denominators. If the
previous value is missing, the engineered change is treated as missing; if the previous value is zero, the
percentage change is set to 0.0 rather than allowing an infinite value. The corresponding ratio features are

\begin{align}
    r_1(x_t) &= \frac{x_t}{x_{t-1}}, \\
    r_2(x_t) &= \frac{x_{t-1}}{x_{t-2}}.
\end{align}

These ratio features are also computed with a safe denominator rule: a missing previous value remains missing,
whereas a zero previous value is assigned a neutral ratio of 1.0. To smooth short-term fluctuations, the model
includes right-aligned rolling means:

\begin{align}
    \mathrm{RM}_{2,t}
    &=
    (x_t + x_{t-1})/2, \\
    \mathrm{RM}_{4,t}
    &=
    (x_t + x_{t-1} + x_{t-2} + x_{t-3})/4.
\end{align}

Because the rolling windows are right-aligned, they use only week $t$ and earlier values. The first difference is,

\begin{equation}
    s_t = x_t - x_{t-1},
\end{equation}

and the second difference is,

\begin{equation}
    a_t = (x_t - x_{t-1}) - (x_{t-1} - x_{t-2}).
\end{equation}

These transformations are applied to every retained exogenous predictor and are also applied separately to the
observed history of the active target. In the single-pathogen paradigm, the active target is the corresponding
disease admission series; for example, an RSV run may use Total RSV Admissions and its lags, percentage
changes, ratios, rolling means, first difference, and second difference, but not Total RSV Admissions T+h as an
input feature. In the multi-pathogen paradigm, the same rule is applied to the observed combined admission
history, but not to Combined Admissions T+h. Therefore, the feature set uses week $t$ and prior information to
predict week $t+h$, while the shifted target is used only as the supervised learning label.

Seasonality is represented by Fourier terms derived from ISO week number:
\begingroup
\setlength{\abovedisplayskip}{8pt}
\setlength{\belowdisplayskip}{4pt}
\setlength{\abovedisplayshortskip}{3pt}
\setlength{\belowdisplayshortskip}{0pt}
\setlength{\jot}{0pt}
\begin{align}
    S_t &=
    \sin\left(\frac{2\pi \cdot \mathrm{weekNum}_t}{52}\right), \\
    C_t &=
    \cos\left(\frac{2\pi \cdot \mathrm{weekNum}_t}{52}\right).
\end{align}
\endgroup

These seasonal terms provide a compact annual cycle without requiring the model to estimate a separate
coefficient for each week of the year. After feature engineering, infinite values are converted to missing
values. The engineered data are then separated into two related objects for each target-horizon pair. For
retrospective training and testing, the labeled modeling frame keeps only rows that have both a complete
engineered predictor vector and a known shifted target for that horizon; rows with missing engineered
predictors or unknown future labels are excluded. This separation keeps the retrospective and prospective
datasets aligned in their predictor definitions while allowing their target-availability requirements to
differ. It also prevents information from future admission weeks from entering the engineered predictor matrix
before forecasts are issued. For prospective forecasting, the
future target is not required because it is the value being predicted. The pipeline therefore also retains the
most recent row with complete engineered predictors as the forecast-origin row, even though its T+1, T+2, T+3,
or T+4 admission value may not yet be observed.

\begin{table}[!htbp]
\centering
\caption{Disease-specific selected and engineered feature counts from the model runs.}
\label{tab:training_summary}
\vspace{0.6em}
\scriptsize
\renewcommand{\arraystretch}{1.0}
\setlength{\tabcolsep}{3pt}
\resizebox{\columnwidth}{!}{%
\begin{tabular}{cccc}
\toprule
Target & Horizon & Selected Features & Engineered Features \\
\midrule
Influenza & T+1 & 20 & 242 \\
Influenza & T+2 & 18 & 230 \\
Influenza & T+3 & 19 & 242 \\
Influenza & T+4 & 17 & 218 \\
COVID-19 & T+1 & 20 & 242 \\
COVID-19 & T+2 & 20 & 242 \\
COVID-19 & T+3 & 18 & 218 \\
COVID-19 & T+4 & 19 & 230 \\
RSV & T+1 & 21 & 254 \\
RSV & T+2 & 19 & 242 \\
RSV & T+3 & 16 & 206 \\
RSV & T+4 & 19 & 242 \\
\bottomrule
\end{tabular}
}
\end{table}

\begin{table}[!htbp]
\centering
\caption{Combined-target selected and engineered feature counts from the model runs.}
\label{tab:combined_training_summary}
\vspace{0.6em}
\scriptsize
\renewcommand{\arraystretch}{1.0}
\setlength{\tabcolsep}{3pt}
\resizebox{\columnwidth}{!}{%
\begin{tabular}{cccc}
\toprule
Target & Horizon & Selected Features & Engineered Features \\
\midrule
Combined & T+1 & 28 & 280 \\
Combined & T+2 & 28 & 280 \\
Combined & T+3 & 28 & 280 \\
Combined & T+4 & 26 & 256 \\
\bottomrule
\end{tabular}
}
\end{table}

\FloatBarrier

\subsubsection{Model Design and Training}

\paragraph*{XGBoost Model}

To isolate the role of forecast-target construction, each target--horizon pair was modeled with a separate
standard XGBoost regressor \citep{Chen2016XGBoost}. The same model form was used for the three
single-pathogen targets and for the combined respiratory-burden target, with separate fitted models for T+1
through T+4. XGBoost is a gradient-boosted tree model for continuous outcomes; here, the outcome is the future
weekly admission count for the selected target and horizon. It was used as an operational supervised
prediction model for weekly respiratory-admission burden, not as a mechanistic model of pathogen transmission.
The primary machine-learning formulation was multivariate supervised time-series prediction: each target was
predicted from its own recent admission history together with contemporaneously available local, clinical, and
surveillance predictors.

For target $q$, week $t$, and horizon $h$, the prediction is the sum of the contributions from all regression
trees:

\begin{equation}
    \hat{y}_{t,h}^{(q)}
    =
    \sum_{m=1}^{M}
    f_m^{(q,h)}(\mathbf{x}_t),
\end{equation}

where $\mathbf{x}_t$ is the engineered feature vector and $f_m^{(q,h)}$ denotes the $m$th regression tree for
target $q$ and horizon $h$. Model settings are selected through the training procedure described below.

\paragraph*{Model Training}

After feature engineering, each target- and horizon-specific labeled dataset was kept in chronological order
and divided into an outer training segment, a temporal gap, and a retrospective test segment:

\begin{equation}
    D_{\mathrm{model}}
    =
    \underbrace{D_{\mathrm{train}}}_{\text{time-series validation inside}}
    \;|\;
    D_{\mathrm{gap}}
    \;|\;
    D_{\mathrm{test}}.
    \raisetag{0.5\baselineskip}
\end{equation}

$D_{\mathrm{train}}$ contains the earlier labeled weeks used for model training.
$D_{\mathrm{train}}$ spanned 2020-09-05 through 2025-10-11. $D_{\mathrm{gap}}$ contains the skipped weeks
between the development and test periods; this gap spanned 2025-10-18 through 2025-11-08. The gap was included
to preserve temporal separation, prevent leakage from observations immediately adjacent to the retrospective
test period, and ensure that model selection used only information available before the held-out evaluation
period. $D_{\mathrm{test}}$ contains later weeks held out for retrospective evaluation. The outer gap and test
rows are not used during model selection.

Within $D_{\mathrm{train}}$, Optuna with the Tree-structured Parzen Estimator sampler is used for
hyperparameter tuning \citep{Akiba2019Optuna,Bergstra2011TPE}. Tuning uses rolling-origin time-series
validation. For validation split $s$, the training segment is divided into an earlier fitting block and a
later validation block:

\begin{equation}
    D_{\mathrm{train}}^{(s)}
    =
    D_{\mathrm{fit}}^{(s)}
    \;|\;
    D_{\mathrm{val}}^{(s)}.
\end{equation}

For that split, the candidate model is trained only on $D_{\mathrm{fit}}^{(s)}$ and scored only on
$D_{\mathrm{val}}^{(s)}$. The validation block follows the fitting block in time, and a validation gap equal
to the forecast horizon separates fitting rows from validation rows. For all diseases and horizons, tuning used
a fixed random seed of 42, an Optuna TPE sampler with seed 42, 25 trials, and 10 rolling-origin validation
splits.
Optuna selected hyperparameters by maximizing validation PA aggregated across the chronological validation
splits. This preserves temporal order and follows forecast evaluation practice emphasizing standardized
horizons and retrospective assessment because forecast accuracy varies over time, geography, model class, and
lead time.
The development workflow therefore involved 16 target--horizon tuning runs: influenza, COVID-19, RSV, and the
combined burden target across four forecast horizons. With 25 trials and 10 rolling-origin splits per run,
hyperparameter search was the main computational cost, whereas prospective operation reused the selected
settings and required only weekly data updating, feature construction, final model fitting, and prediction.
This separation limited issue-time latency and clarifies the distinction between model development and
real-time forecast generation \citep{Pollett2021EPIFORGE}.

For each target and horizon, the search space $\Theta$ includes n estimators (tree count) from 200 to 800,
max depth (tree depth) from 3 to 8, learning rate ($\eta$; step size) from 0.01 to 0.10, subsample (row sampling)
and colsample bytree (feature sampling) from 0.60 to 1.00, min child weight (split size) from 1 to 10,
gamma (split penalty) from 0 to 1, reg alpha ($\alpha$; L1 penalty) from 0 to 1, and reg lambda
($\lambda$; L2 penalty) from
0 to 5. The Optuna-selected values are summarized for the disease-specific runs in
Table~3 and for the combined-burden model in Table~4.

\begin{NoHyper}
\begin{table*}[!tbp]
\centering
\caption{Optuna-selected XGBoost hyperparameters by disease and forecast horizon from the model runs.}
\label{tab:best_hyperparameters}
\scriptsize
\renewcommand{\arraystretch}{0.95}
\setlength{\tabcolsep}{3pt}
\begin{tabular*}{\textwidth}{@{\extracolsep{\fill}}ccccccccccc@{}}
\toprule
Target & Horizon & Trees & Depth & $\eta$ & Subsample & Colsample & Child & Gamma & $\alpha$ & $\lambda$ \\
\midrule
Influenza & T+1 & 414 & 4 & 0.0588 & 0.656 & 0.921 & 1 & 0.987 & 0.772 & 0.994 \\
Influenza & T+2 & 537 & 7 & 0.0666 & 0.828 & 0.707 & 1 & 0.896 & 0.741 & 3.755 \\
Influenza & T+3 & 383 & 3 & 0.0716 & 0.776 & 0.649 & 5 & 0.034 & 0.909 & 1.294 \\
Influenza & T+4 & 256 & 7 & 0.0283 & 0.708 & 0.851 & 3 & 0.564 & 0.396 & 0.552 \\
COVID-19 & T+1 & 786 & 3 & 0.0432 & 0.949 & 0.751 & 3 & 0.222 & 0.010 & 0.368 \\
COVID-19 & T+2 & 286 & 3 & 0.0674 & 0.779 & 0.905 & 5 & 0.226 & 0.791 & 3.333 \\
COVID-19 & T+3 & 517 & 5 & 0.0701 & 0.874 & 0.640 & 7 & 0.272 & 0.701 & 1.218 \\
COVID-19 & T+4 & 707 & 3 & 0.0951 & 0.969 & 0.774 & 8 & 0.227 & 0.427 & 2.883 \\
RSV & T+1 & 675 & 8 & 0.0801 & 0.877 & 0.643 & 2 & 0.857 & 0.997 & 1.922 \\
RSV & T+2 & 509 & 6 & 0.0142 & 0.843 & 0.668 & 1 & 0.949 & 0.966 & 4.042 \\
RSV & T+3 & 696 & 6 & 0.0286 & 0.875 & 0.876 & 9 & 0.576 & 0.036 & 3.356 \\
RSV & T+4 & 668 & 5 & 0.0267 & 0.998 & 0.852 & 10 & 0.592 & 0.432 & 3.525 \\
\bottomrule
\end{tabular*}

\par\vspace{0.75em}

\caption{Optuna-selected XGBoost hyperparameters for the combined respiratory-burden model.}
\label{tab:combined_best_hyperparameters}
\scriptsize
\renewcommand{\arraystretch}{0.95}
\setlength{\tabcolsep}{3pt}
\begin{tabular*}{\textwidth}{@{\extracolsep{\fill}}ccccccccccc@{}}
\toprule
Target & Horizon & Trees & Depth & $\eta$ & Subsample & Colsample & Child & Gamma & $\alpha$ & $\lambda$ \\
\midrule
Combined & T+1 & 349 & 8 & 0.0889 & 0.869 & 0.866 & 4 & 0.504 & 0.454 & 1.806 \\
Combined & T+2 & 305 & 5 & 0.0496 & 0.608 & 0.956 & 4 & 0.944 & 0.607 & 0.639 \\
Combined & T+3 & 326 & 3 & 0.0461 & 0.617 & 0.756 & 6 & 0.242 & 0.454 & 0.136 \\
Combined & T+4 & 294 & 4 & 0.0516 & 0.804 & 0.728 & 10 & 0.559 & 0.456 & 1.948 \\
\bottomrule
\end{tabular*}
\end{table*}
\end{NoHyper}

For a trial parameter set $\theta \in \Theta$, predictions are made for the validation rows across the
chronological splits and scored using the MAE and PA definitions in
Section~\ref{sec:performance_evaluation}. The same nonnegative clipping rule defined in
Section~\ref{sec:performance_evaluation} was applied before validation scoring and final evaluation. Let
$\mathcal{T}_h \subset \Theta$ denote the configurations evaluated by Optuna for horizon $h$. The selected
parameter set is the trial that maximizes validation PA.
\vspace{1 em}

After tuning hyperparameters, the selected parameter set is used to fit a retrospective test model on the training segment for
retrospective evaluation. For each forecasting target and horizon, the retrospective test model is trained using only
$D_{\mathrm{train}}$. The gap rows are skipped, so they are not used for fitting:

\begin{equation}
    F^{\mathrm{test}}_h
    =
    \operatorname{Train}_h(D_{\mathrm{train}}).
\end{equation}

For prospective forecasting, hyperparameters are not re-tuned.
The selected parameters are carried forward to fit a final model on all complete rows with known
horizon-specific labels before the forecast origin:

\begin{equation}
\begin{aligned}
    \mathcal{D}_{\mathrm{final},h}
    =
    \{(\mathbf{x}_t, y_{t,h}) :&
    \mathbf{x}_t \ \mathrm{complete}, \\
    & y_{t,h} \ \mathrm{known}, \\
    & t < t_{\mathrm{origin}}\}.
\end{aligned}
\end{equation}

\begin{equation}
    F^{\mathrm{final}}_h
    =
    \operatorname{Train}_h(\mathcal{D}_{\mathrm{final},h}).
\end{equation}

This final model is applied to the latest complete feature row to generate the T+1, T+2, T+3, and T+4 point
predictions.

\subsection{Adaptive Prediction Generation}

For both retrospective estimation and prospective forecasting, the fitted XGBoost model first generates a raw
prediction. For target $q$, horizon $h$, and feature vector $\mathbf{x}_t$, the raw prediction is

\begin{equation}
    \hat{y}^{\mathrm{raw}}_{q,t,h}
    =
    F_{q,h}(\mathbf{x}_t).
\end{equation}

For retrospective evaluation, raw predictions are generated for the held-out test weeks. For prospective
forecasting, the same raw-prediction form is generated from the final horizon-specific model at the forecast
origin. In both settings, the reported and scored value is the adaptive prediction rather than the unadjusted
raw output. The adjustment uses only historical scored weeks available before the prediction being evaluated;
held-out retrospective targets and future prospective targets are not used to estimate the adjustment before
prediction. The historical actual-to-predicted ratio for target $q$ and horizon $h$ is

\begin{equation}
    \rho_{q,h}
    =
    \frac{
    \frac{1}{|\mathcal{H}_{q,h}|}
    \sum_{t \in \mathcal{H}_{q,h}} y_{q,t,h}
    }{
    \frac{1}{|\mathcal{H}_{q,h}|}
    \sum_{t \in \mathcal{H}_{q,h}} \hat{y}^{\mathrm{raw}}_{q,t,h}
    },
\end{equation}

where $\mathcal{H}_{q,h}$ denotes the historical scored weeks used for the target- and horizon-specific adjustment. The
adaptive prediction is then generated as

\begin{equation}
    \hat{y}^{\mathrm{adapt}}_{q,t,h}
    =
    \max
    \left(
    \hat{y}^{\mathrm{raw}}_{q,t,h}
    \cdot
    \rho_{q,h}
    \cdot
    \alpha_{q,h},
    0
    \right),
\end{equation}

where $\alpha_{q,h}$ is a target- and horizon-specific smoothing constant with $0<\alpha_{q,h}<1$, chosen during
retrospective workflow development based on PA and then fixed for retrospective and prospective scoring. Thus,
the adaptive prediction rescales the raw output by $\rho_{q,h}$ and $\alpha_{q,h}$.

\subsection{Baseline Models}

Baseline comparisons were defined to contextualize both the disease-specific models and the combined-burden
model before interpreting the adaptive forecasts. These comparisons show how the raw XGBoost model performs
relative to standard time-series and tree-ensemble references. For the combined target, the same comparator set
evaluates the combined-burden XGBoost output relative to alternative model classes. Baseline or reference models are commonly used in infectious disease forecast evaluations to determine
whether a proposed approach improves on simpler forecasting strategies and to make performance differences
easier to interpret across horizons and settings \citep{Reich2019InfluenzaAssessment,Cramer2022COVIDHub}.
Baseline models were fit separately for influenza, COVID-19, RSV, and the combined target at each forecast
horizon. For target $q \in \{\mathrm{Influenza}, \mathrm{COVID\mbox{-}19}, \mathrm{RSV},
\mathrm{Combined}\}$ and horizon $h$, the comparison included Baseline XGBoost, SARIMA, Random Forest, and the
adaptive XGBoost model.

Baseline XGBoost was defined as the raw XGBoost prediction:

\begin{equation}
    \hat{y}_{t,h}^{\mathrm{base\mbox{-}XGB},(q)}
    =
    F_h^{(q)}(\mathbf{x}_t).
\end{equation}

SARIMA was retained as a univariate seasonal time-series baseline. For each target, it forecasts future
admissions from the historical target series rather than from the full engineered feature set:

\begin{equation}
    \hat{y}_{t,h}^{\mathrm{SARIMA},(q)}
    =
    S_h^{(q)}(y_t^{(q)}, y_{t-1}^{(q)}, \ldots).
\end{equation}

Random Forest was used as a non-boosted tree-ensemble baseline. It uses the same horizon-specific feature
matrix as the machine-learning models and averages predictions across regression trees:

\begin{equation}
    \hat{y}_{t,h}^{\mathrm{RF},(q)}
    =
    \frac{1}{B}
    \sum_{b=1}^{B}
    T_b^{(q,h)}(\mathbf{x}_t).
\end{equation}

Together, these baselines compare the adaptive XGBoost forecasts with an unadjusted boosting model, a seasonal
time-series model, and a standard tree-ensemble model for the target-specific forecasts. The disease-specific
and combined-burden baseline results are reported in Tables~\ref{tab:performance} and
\ref{tab:combined_baseline_performance}.

\section{Results}

This section reports forecast performance in three stages: retrospective disease-specific results for influenza,
COVID-19, and RSV; retrospective comparison of summed disease-specific forecasts with the
combined-burden model; and prospective comparison of the same total-burden forecasting paradigms.

\subsection{Retrospective Estimation}

Retrospective out-of-sample performance was evaluated after the training and temporal-gap periods for held-out
target weeks from 2025-11-15 through 2026-02-21 across T+1 through T+4 horizons, using the split shown in
Figure~\ref{fig:data_splitting_timeline}. Results are reported first for disease-specific influenza, COVID-19,
and RSV forecasts (Table~\ref{tab:performance}) and the combined-burden model
(Table~\ref{tab:combined_baseline_performance}), followed by the total-burden comparison between summed
disease-specific forecasts and the combined-burden model (Tables~\ref{tab:summed_single_pathogen_performance}
and \ref{tab:combined_performance}; Figure~\ref{fig:retrospective_combined_admissions_pa_mae}).
This ordering separates disease-level prediction performance from total-burden estimation, making the
single-pathogen and combined-target results interpretable before the two burden paradigms are compared.

\clearpage
\onecolumn

\begin{center}
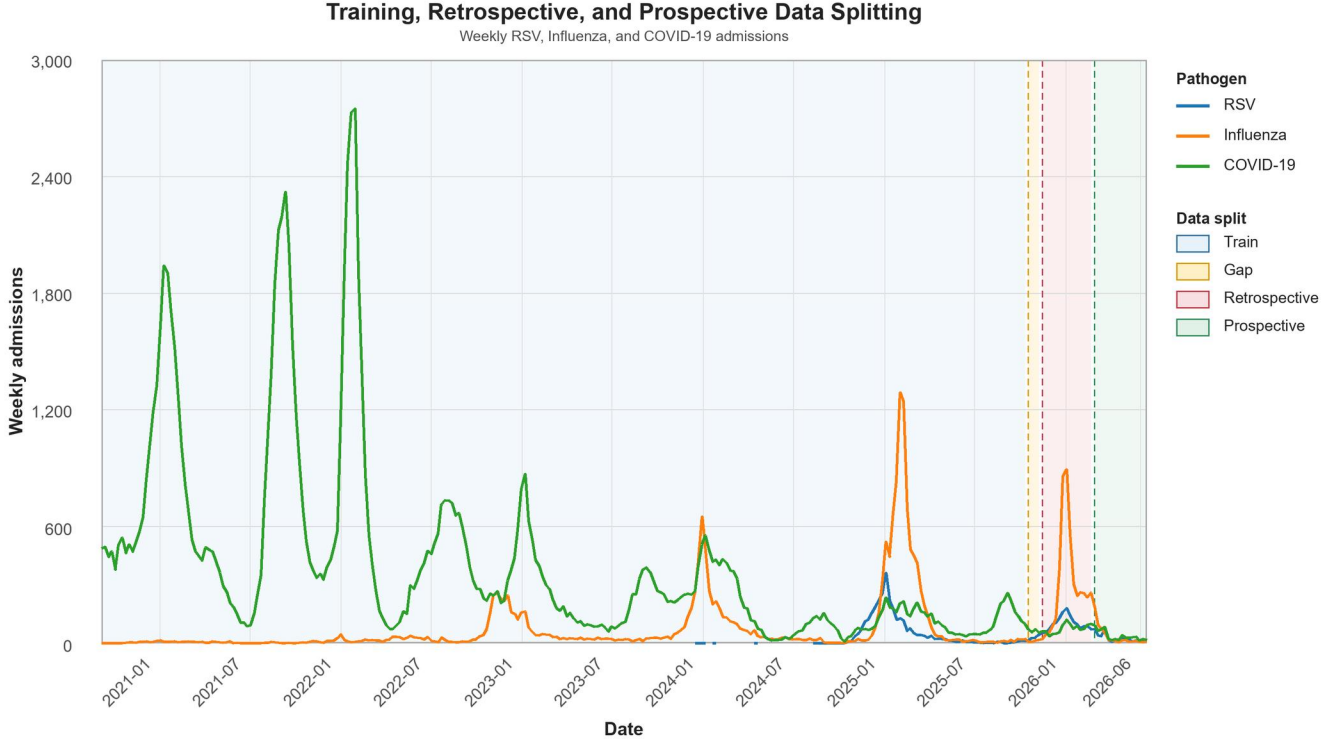

\centering
\setlength{\abovecaptionskip}{2pt}
\setlength{\belowcaptionskip}{3pt}
\vspace*{-0.7em}
\DataSplittingTimelineFigure
\captionof{figure}{Training, gap, retrospective, and prospective data split used for respiratory-admission forecasting.}
\label{fig:data_splitting_timeline}
\end{center}

\begingroup
\vspace{-0.9em}
\begin{center}
\centering
\captionof{table}{Retrospective results across disease-specific models.}
\label{tab:performance}
\vspace{0.35em}
\scriptsize
\renewcommand{\arraystretch}{1.0}
\setlength{\tabcolsep}{3pt}
\resizebox{\textwidth}{!}{%
\begin{tabular}{@{}cccccccccc@{}}
\toprule
Disease & Horizon &
\shortstack{Baseline XGBoost\\MAE} &
\shortstack{SARIMA\\MAE} &
\shortstack{Random Forest\\MAE} &
\shortstack{Adaptive XGBoost\\MAE} &
\shortstack{Baseline XGBoost\\PA} &
\shortstack{SARIMA\\PA} &
\shortstack{Random Forest\\PA} &
\shortstack{Adaptive XGBoost\\PA} \\
\midrule
Influenza & T+1 & 88.84 & 92.01  & 57.16  & 74.64 & 0.80 &0.75  & 0.83 & 0.84 \\
Influenza & T+2 & 190.88 &209.70  &142.90  & 189.58 & 0.77 &0.51  &0.69  & 0.75 \\
Influenza & T+3 & 164.33 &286.84  &152.86  & 176.53 & 0.63 &0.40  &0.69  & 0.62 \\
Influenza & T+4 & 194.53 & 338.12 &158.58  & 191.29 & 0.58 & 0.28  &0.67  & 0.62 \\
COVID-19 & T+1 & 18.66 & 26.00 &18.39  & 20.70 & 0.79 &0.66  & 0.78 & 0.75 \\
COVID-19 & T+2 & 21.53 &34.48  &29.53  & 17.92 & 0.76 &0.57  & 0.71 & 0.80 \\
COVID-19 & T+3 & 23.41 & 37.31 &  32.73 & 24.88 & 0.73 &0.54  &0.70  & 0.73 \\
COVID-19 & T+4 & 27.51 & 36.80 & 36.15 & 19.60 & 0.69 &0.54  &0.67  & 0.79 \\
RSV & T+1 & 17.18 & 38.71 & 37.84  & 11.67 & 0.81 &0.65  &0.69  & 0.89 \\
RSV & T+2 & 29.83 &40.54  & 38.05 & 21.57 & 0.75 & 0.65 &0.70  & 0.81 \\
RSV & T+3 & 30.98 &35.02  &38.91  & 25.64 & 0.70 & 0.68 &0.69  & 0.80 \\
RSV & T+4 & 21.20 &89.73  & 33.13 & 26.85 & 0.67 & 0.53 & 0.73 & 0.79 \\
\bottomrule
\end{tabular}
}
\end{center}
\endgroup

\vspace{-0.4em}
\begingroup
\begin{center}
\centering
\captionof{table}{Retrospective results across combined-burden models.}
\label{tab:combined_baseline_performance}
\scriptsize
\renewcommand{\arraystretch}{1.0}
\setlength{\tabcolsep}{3pt}
\resizebox{\textwidth}{!}{%
\begin{tabular}{@{}cccccccccc@{}}
\toprule
Disease & Horizon &
\shortstack{Baseline XGBoost\\MAE} &
\shortstack{SARIMA\\MAE} &
\shortstack{Random Forest\\MAE} &
\shortstack{Adaptive XGBoost\\MAE} &
\shortstack{Baseline XGBoost\\PA} &
\shortstack{SARIMA\\PA} &
\shortstack{Random Forest\\PA} &
\shortstack{Adaptive XGBoost\\PA} \\
\midrule
Combined & T+1 & 114.42 & 144.52 & 114.32 & 80.63  & 0.83 & 0.73 & 0.81 & 0.85 \\
Combined & T+2 & 228.97 & 251.86 & 216.66 & 200.09 & 0.67 & 0.76 & 0.78 & 0.78 \\
Combined & T+3 & 289.41 & 257.83 & 160.32 & 183.97 & 0.62 & 0.72 & 0.74 & 0.74 \\
Combined & T+4 & 291.27 & 140.57 & 164.56 & 156.13 & 0.61 & 0.74 & 0.74 & 0.76 \\
\bottomrule
\end{tabular}
}
\end{center}
\endgroup

\switchtwocolumn

\vspace{1 em}

\begingroup
\setlength{\parskip}{0pt}

Table~\ref{tab:performance} shows disease- and horizon-specific variation across comparator models. For
influenza, Adaptive XGBoost PA was 0.84, 0.75, 0.62, and 0.62 from T+1 to T+4, with MAE values of 74.64,
189.58, 176.53, and 191.29. COVID-19 PA was 0.75, 0.80, 0.73, and 0.79, with MAE values of 20.70, 17.92,
24.88, and 19.60. RSV had PA values of 0.89, 0.81, 0.80, and 0.79 and MAE values of 11.67, 21.57, 25.64, and
26.85. Across Baseline XGBoost, SARIMA, Random Forest, and Adaptive XGBoost, no single method dominated both PA
and MAE for every disease--horizon combination.

\noindent Table~\ref{tab:combined_baseline_performance} shows the same comparator-model evaluation for the
combined-burden target. Adaptive XGBoost reduced MAE relative to baseline XGBoost at all horizons, from 114.42
to 80.63 at T+1, 228.97 to 200.09 at T+2, 289.41 to 183.97 at T+3, and 291.27 to 156.13 at T+4. PA also
increased from 0.83, 0.67, 0.62, and 0.61 under baseline XGBoost to 0.85, 0.78, 0.74, and 0.76 after adaptive
adjustment. Among the comparator models, Random Forest had the lowest MAE at T+3 (160.32), while Random Forest
and Adaptive XGBoost had the same PA at that horizon (0.74). Adaptive XGBoost improved MAE and PA relative to
baseline XGBoost at all combined-target horizons, although Random Forest matched or exceeded it for selected
T+3 metrics.

For the total-burden comparison, ``Summed'' refers to the post-modeling sum of the separate influenza,
COVID-19, and RSV adaptive forecasts, while ``Combined'' refers to the model trained directly on the aligned
total respiratory-admission target. Tables~\ref{tab:summed_single_pathogen_performance} and
\ref{tab:combined_performance} report the retrospective total-burden scores, and
Figure~\ref{fig:retrospective_combined_admissions_pa_mae} displays the same comparison across T+1 through T+4.

\endgroup

\vspace{- 1em}
\begin{table}[H]
\centering
\caption{Retrospective results for summed disease-specific forecasts.}
\label{tab:summed_single_pathogen_performance}
\vspace{0.01em}
\scriptsize
\renewcommand{\arraystretch}{1.0}
\setlength{\tabcolsep}{3pt}
\resizebox{\columnwidth}{!}{%
\begin{tabular}{cccc}
\toprule
Target & Horizon &
\shortstack{Adaptive XGBoost\\MAE} &
\shortstack{Adaptive XGBoost\\PA} \\
\midrule
Summed & T+1 & 100.14 & 0.85 \\
Summed & T+2 & 200.06 & 0.80 \\
Summed & T+3 & 205.65 & 0.69 \\
Summed & T+4 & 217.57 & 0.69 \\
\bottomrule
\end{tabular}
}
\end{table}

\vspace{-1.3em}
\noindent\begin{minipage}{\columnwidth}
\begingroup
\centering
\captionof{table}{Retrospective results for the combined-burden model.}
\label{tab:combined_performance}
\vspace{0em}
\scriptsize
\renewcommand{\arraystretch}{1.0}
\setlength{\tabcolsep}{3pt}
\resizebox{\columnwidth}{!}{%
\begin{tabular}{cccc}
\toprule
Target & Horizon &
\shortstack{Adaptive XGBoost\\MAE} &
\shortstack{Adaptive XGBoost\\PA} \\
\midrule
Combined & T+1 & 80.63 & 0.85 \\
Combined & T+2 & 200.09 & 0.78 \\
Combined & T+3 & 183.97 & 0.74 \\
Combined & T+4 & 156.13 & 0.76 \\
\bottomrule
\end{tabular}
}
\endgroup
\par\vspace{0.7 em}
\normalsize
\hspace*{1.5em}Tables~\ref{tab:summed_single_pathogen_performance} and \ref{tab:combined_performance} show that the two
total-burden paradigms were similar at T+1 and T+2 but separated at longer horizons. From T+1 through T+4, the
summed forecasts had MAE values of 100.14, 200.06, 205.65, and 217.57, with PA values of 0.85, 0.80, 0.69, and
0.69. The combined-burden model had MAE values of 80.63, 200.09, 183.97, and 156.13, with PA values of 0.85, 0.78,
0.74, and 0.76. PA was identical at T+1, slightly higher for the summed forecasts at T+2, and higher for the
combined-burden model at T+3 and T+4. The combined-burden model had lower MAE at T+1, T+3, and T+4, while T+2 MAE was nearly
identical between the two approaches.
Figure~\ref{fig:retrospective_combined_admissions_pa_mae} highlights the longer-horizon separation: at T+3 and
T+4, combined PA was 0.74 and 0.76 versus 0.69 and 0.69 for the summed forecasts, with MAE of 183.97 and
156.13 versus 205.65 and 217.57.
\end{minipage}

\FloatBarrier
\par\vspace{0.5em}
\begin{strip}
\centering
\setlength{\abovecaptionskip}{2pt}
\setlength{\belowcaptionskip}{3pt}
\CombinedAdmissionsPaMaeFigure
\captionof{figure}{Retrospective PA and MAE comparison between summed separate single-pathogen forecasts and the
combined-burden model.}
\label{fig:retrospective_combined_admissions_pa_mae}
\end{strip}
\FloatBarrier

\clearpage

\makeatletter
\if@twocolumn\else\switchtwocolumn\fi
\makeatother

\subsection{Prospective Forecasting}

The prospective evaluation assessed operational forecasts issued at the Center for Public Health Modeling and
Response before the target weeks were observed and scored after admission counts became
available. Target timestamps ran from 2026-02-28 through 2026-06-13. The horizon-specific models and calibration
settings selected during retrospective development were carried forward without re-tuning. As in the retrospective
analysis, total burden was scored using two paradigms: summed disease-specific forecasts and the
combined-burden model. Tables~\ref{tab:prospective_summed_single_pathogen_performance} and
\ref{tab:prospective_combined_performance} report the prospective total-burden scores, and
Figure~\ref{fig:prospective_combined_admissions_pa_mae} displays the comparison across T+1 through T+4.

\vspace{-0.8em}
\begin{table}[H]
\centering
\caption{Prospective results for summed disease-specific forecasts.}
\label{tab:prospective_summed_single_pathogen_performance}
\vspace{0.01em}
\scriptsize
\renewcommand{\arraystretch}{1.0}
\setlength{\tabcolsep}{3pt}
\resizebox{\columnwidth}{!}{%
\begin{tabular}{cccc}
\toprule
Target & Horizon &
\shortstack{Adaptive XGBoost\\MAE} &
\shortstack{Adaptive XGBoost\\PA} \\
\midrule
Summed & T+1 & 40.21 & 0.70 \\
Summed & T+2 & 42.46 & 0.69 \\
Summed & T+3 & 64.43 & 0.62 \\
Summed & T+4 & 43.47 & 0.66 \\
\bottomrule
\end{tabular}
}
\end{table}

\vspace{-2.5em}
\begin{table}[H]
\centering
\caption{Prospective results for the combined-burden model.}
\label{tab:prospective_combined_performance}
\vspace{0.2em}
\scriptsize
\renewcommand{\arraystretch}{1.0}
\setlength{\tabcolsep}{3pt}
\resizebox{\columnwidth}{!}{%
\begin{tabular}{cccc}
\toprule
Target & Horizon &
\shortstack{Adaptive XGBoost\\MAE} &
\shortstack{Adaptive XGBoost\\PA} \\
\midrule
Combined & T+1 & 36.33 & 0.71 \\
Combined & T+2 & 37.57 & 0.70 \\
Combined & T+3 & 33.19 & 0.70 \\
Combined & T+4 & 21.97 & 0.76 \\
\bottomrule
\end{tabular}
}
\end{table}
\vspace{-1.2em}

Tables~\ref{tab:prospective_summed_single_pathogen_performance} and
\ref{tab:prospective_combined_performance} show that the combined-burden model had higher PA and lower MAE than the
summed forecasts at all four prospective horizons. From T+1 through T+4, the summed forecasts had MAE values of
40.21, 42.46, 64.43, and 43.47, with PA values of 0.70, 0.69, 0.62, and 0.66. The combined-burden model had MAE values
of 36.33, 37.57, 33.19, and 21.97, with PA values of 0.71, 0.70, 0.70, and 0.76. Differences were small at T+1
and T+2 and larger at T+3 and T+4; the largest MAE separation occurred at T+3, while the largest PA separation
occurred at T+4.
Figure~\ref{fig:prospective_combined_admissions_pa_mae} shows the same prospective pattern visually: at T+3 and
T+4, combined MAE was 33.19 and 21.97 versus 64.43 and 43.47 for the summed forecasts, with PA of 0.70 and
0.76 versus 0.62 and 0.66.
\FloatBarrier
\par\vspace{0.5em}

\begin{strip}
\centering
\setlength{\abovecaptionskip}{2pt}
\setlength{\belowcaptionskip}{3pt}
\CombinedAdmissionsProspectivePaMaeFigure

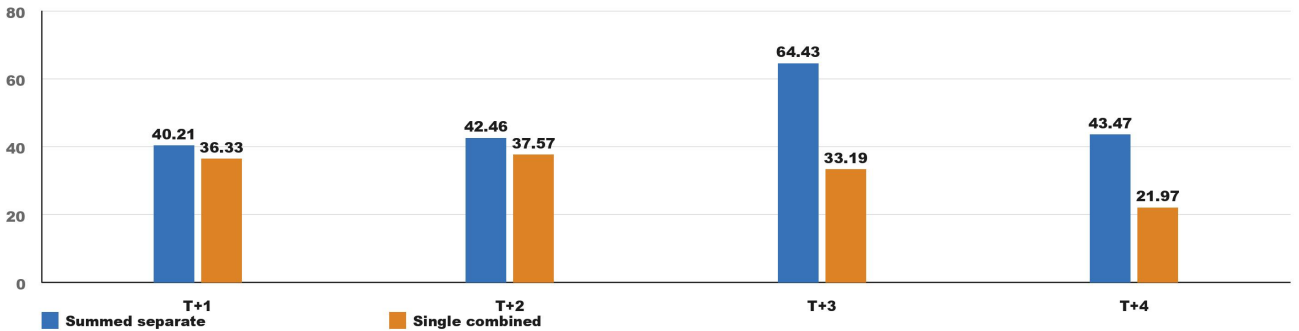
\captionof{figure}{Prospective PA and MAE comparison between summed separate single-pathogen forecasts and the
combined-burden model.}
\label{fig:prospective_combined_admissions_pa_mae}
\end{strip}
\FloatBarrier

\clearpage
\section{Discussion}

This study evaluated a central operational forecasting question: whether short-term respiratory hospital burden is
better estimated with a unified multi-pathogen forecasting framework or by independently developed
pathogen-specific models. The results support the unified combined-burden approach when the endpoint is total
admissions requiring beds, staff, testing capacity, and respiratory support. Disease-specific forecasts still
characterize influenza, COVID-19, and RSV trajectories separately, but the operational total-burden signal is
better treated as its own forecasting target.

This distinction is more than a technical modeling choice. Forecast target definition determines what error the
model is trained to minimize and what operational question the forecast is designed to answer. When independently
developed pathogen-specific forecasts are later aggregated, the resulting total inherits the timing, calibration,
and scale errors of three separately tuned models. A combined-burden forecast instead learns the aggregate
admission trajectory directly as its own target. This study therefore contributes to the growing
literature on real-time and multi-source respiratory forecasting by focusing on target construction as well as
model class
\citep{Mellor2025,Ray2025Flusion,Meyer2025TransferLearning,Xu2025OGEM}. The combined-burden target should
therefore be interpreted as a decision-relevant admission signal, not as a mechanistic representation of
influenza, COVID-19, and RSV transmission.

\subsection{Retrospective Estimation Analysis}

The retrospective evaluation showed that forecast behavior differed by pathogen, forecast horizon, and metric
(Table~\ref{tab:performance}). RSV had the most stable disease-specific agreement across horizons, whereas
influenza showed the clearest deterioration at longer horizons. COVID-19 did not follow a simple monotonic
pattern, which is consistent with the broader challenge of forecasting a pathogen whose recent hospitalization
signal has been shaped by changing immunity, testing, reporting, variants, and public behavior
\citep{Klein2023,Lopez2024COVIDCases}. These differences matter because the aggregated pathogen-specific
paradigm, when used for total-burden planning, is only as stable as its separate components. If one pathogen has a
timing error, another has a scale error, and a third has more stable relative agreement, the aggregate forecast
can improve through accidental cancellation or deteriorate through error accumulation.

The comparison with baseline XGBoost, SARIMA, and Random Forest also shows why a single accuracy measure is not
sufficient. Adaptive XGBoost was strongest for RSV agreement across the four horizons and improved selected
COVID-19 horizons, whereas Random Forest produced lower influenza MAE at several horizons. This absence of a
universal winner is common in epidemic forecasting, where models may differ in short-term tracking, peak timing,
calibration, and scale sensitivity \citep{Reich2019InfluenzaAssessment,Cramer2022COVIDHub}. PA and MAE therefore
serve complementary roles. PA describes proportional agreement between predicted and observed
admissions, while MAE describes the average number of admissions missed. For public health use, both are needed:
a model that follows the trajectory but misses the admission count can still mislead staffing decisions, while a
model with low average error may still fail to track the timing of a surge.

At the total-burden level, the retrospective results indicate that the aggregated pathogen-specific and
combined-burden paradigms were most similar at the shortest horizons and separated more clearly at longer horizons
(Tables~\ref{tab:summed_single_pathogen_performance} and \ref{tab:combined_performance};
Figure~\ref{fig:retrospective_combined_admissions_pa_mae}). The key interpretation is not simply that the
combined-burden model had better scores at horizon three and horizon four. Figure~\ref{fig:retrospective_combined_admissions_pa_mae} makes this separation visible across both PA and MAE panels. Rather, the longer-horizon pattern suggests that the
combined target preserved the collective respiratory admission signal more effectively when disease-specific
forecasts became harder to synchronize. This is plausible because the combined-burden model does not require each
pathogen-specific model to correctly estimate its own timing and magnitude before aggregation. It directly
optimizes the quantity that is later evaluated.

This interpretation should be treated carefully. The present analysis does not prove that error accumulation is
the sole mechanism behind the summed model's weaker longer-horizon performance. Differences may also reflect the
reconstructed RSV history, the selected features, horizon-specific calibration, the timing of respiratory waves,
or the particular evaluation period. Still, the result is operationally meaningful because horizon three and horizon four forecasts
are often the horizons at which hospitals make schedule, staffing, supply, and capacity decisions that cannot be
changed immediately.

\subsection{Prospective Performance Analysis}

The prospective evaluation is especially important because it tested the workflow under real-time conditions
rather than only through retrospective simulation. Forecasts were issued before the target weeks were observed,
using the model choices and calibration settings carried forward from retrospective development. This mirrors the
operational setting emphasized in recent respiratory forecasting work, where timeliness, repeatability, and
decision relevance are as important as retrospective fit \citep{Mellor2025,Meyer2025TransferLearning,Ahammed2025LocalHealthSystem}.
This prospective step therefore tested the aggregated pathogen-specific versus unified combined-burden comparison under the timing constraints that would
govern routine forecast use.

The prospective results showed a more consistent advantage for the combined-burden model than the retrospective results
(Tables~\ref{tab:prospective_summed_single_pathogen_performance} and
\ref{tab:prospective_combined_performance}; Figure~\ref{fig:prospective_combined_admissions_pa_mae}). The
combined-burden model had higher PA and lower MAE at every horizon, with the largest separation at T+3 and T+4. Figure~\ref{fig:prospective_combined_admissions_pa_mae} shows the same prospective pattern visually, with clearer separation at the longer horizons. This
pattern strengthens the practical argument for using a unified burden target rather than aggregating separate
disease forecasts when the operational target is total respiratory burden: in real-time use, the combined target
remained competitive at the shortest horizons and became more favorable at the horizons most relevant for advance
planning.

The prospective MAE values should not be interpreted as a simple function of forecast horizon. Absolute error
also depends on the level and shape of admissions during the evaluated weeks, and the prospective period covered a
different calendar interval than the retrospective analysis. For that reason, the most defensible comparison is
within the same evaluation period and horizon. Under that comparison, the combined-burden model had more
favorable point-forecast performance.

\subsection{Public Health Relevance}

The public health relevance of this work is that it separates two operational needs that are often conflated.
Disease-specific forecasts for influenza, COVID-19, and RSV help identify which pathogen is contributing to
expected admissions, while the combined-burden forecast estimates the total respiratory pressure hospitals must
manage. In this study, the combined-burden model provided the more consistent total-admission signal, especially
at longer planning horizons.

This distinction matters because staffing, bed management, testing capacity, respiratory-support planning, and
regional coordination depend on total burden rather than pathogen-specific counts alone
\citep{Bi2025,Mellor2025,Brouwer2026Practice}. The framework therefore preserves pathogen-specific
interpretation while adding a hospital-facing burden estimate for respiratory-season preparedness, particularly
when influenza, COVID-19, and RSV overlap or peak at different times \citep{Bi2025,Dewey2025,Mellor2025}.

The 1--4 week horizon also has practical meaning. One-week forecasts can support immediate situational awareness
and short-term staffing adjustments. Forecasts three and four weeks ahead are less certain but more useful for actions
that require lead time, such as schedule planning, supply ordering, surge protocols, and coordination with public
health partners. The finding that the combined-burden model had more favorable PA and MAE at these longer
planning horizons therefore has direct relevance for preparedness, even though the present analysis does not yet
translate forecast errors into specific operational cost or capacity outcomes.

\subsection{Future Extensions}

Several limitations define the next steps for this work. First, the evaluation used data from one geographic
setting and a limited number of respiratory seasons. Generalizability should therefore be tested across additional
states, health systems, hospital catchment areas, and reporting environments where population structure, health
system coverage, testing practice, and admission coding may differ \citep{Ahammed2025LocalHealthSystem}. Second,
the analysis used point forecasts. This is useful for comparing PA and MAE, but operational decision-making also
requires calibrated uncertainty intervals, especially when forecasts are used to prepare for high-consequence
surges \citep{Pollett2021EPIFORGE,Brouwer2026Practice}. Third, the evaluation did not formally quantify whether
score differences were statistically distinguishable across horizons, nor did it decompose summed-forecast error
into pathogen-specific timing, magnitude, and calibration components.

Future work should extend the framework in five directions. Longer prospective evaluation is needed to determine
whether the combined-burden model's advantage persists across multiple respiratory seasons, off-season activity, repeated
waves, and high-burden peaks. Probabilistic forecasts should be added so that hospitals can reason about plausible
ranges rather than single values. Coherent hierarchical forecasting should be explored so that pathogen-specific
forecasts and total-burden forecasts can be reconciled rather than treated as competing products. Additional data
streams, including wastewater, syndromic, mobility, laboratory, and local health-system signals, may improve
timeliness and robustness when traditional reporting is delayed or revised
\citep{Ray2025Flusion,Wu2025Transformer,Tabassum2026Wastewater}. Finally, the forecasts should be linked to
decision thresholds, such as expected bed demand, staffing triggers, testing-capacity thresholds, or respiratory
support readiness, so forecast performance can be evaluated in terms of actionability rather than accuracy alone.

\subsection{Conclusion}

This study suggests that respiratory-hospital forecasting can benefit from treating total respiratory burden as a
distinct forecasting target when the planning question concerns shared hospital capacity. Disease-specific
forecasts provide the pathogen-level information needed to monitor influenza, COVID-19, and RSV separately and
interpret burden composition. In this evaluation, however, the combined-burden model more consistently captured
shared respiratory admission demand than the total obtained from separate disease-specific forecasts, with the
difference most apparent at longer planning horizons.

These findings support a two-output forecasting framework: disease-specific models identify respiratory-burden
composition, while the combined-burden model estimates shared capacity pressure. Combined-burden forecasting may
therefore provide a decision-aligned endpoint for respiratory-season planning and a scalable foundation for
epidemic and pandemic preparedness when multiple pathogens create overlapping hospital pressure.

\section*{CRediT Authorship Contribution Statement}

Gaurob Saha: Conceptualization, methodology, software, validation, formal analysis, investigation,
visualization, project administration, writing -- original draft, writing -- review and editing. Minjae Woo:
Writing -- review and editing. Lior Rennert: Writing -- review and editing.

\section*{Ethics and Data Governance Statement}

This study used aggregate or de-identified respiratory-surveillance and hospital-utilization data. No
individual-level identifiable information is reported. Data use followed the applicable institutional and
data-provider approvals and restrictions.

\section*{Declaration of Competing Interest}

The authors declare that they have no known competing financial interests or personal relationships that could
have appeared to influence the work reported in this paper.

\section*{Funding}

The authors acknowledge the Public Health Science department at Clemson University for providing facilities and support. 

\section*{Acknowledgements}

The authors acknowledge the data providers and collaborators supporting respiratory disease surveillance and
forecasting.

\section*{Data Availability}

The complete analytic dataset is not publicly available because it combines public CDC NHSN source data with
restricted South Carolina RFA hospitalization data and internal MUSC and PRISMA surveillance indicators. Public
CDC NHSN source components can be obtained from CDC data repositories. RFA hospitalization data may be available
from South Carolina RFA subject to approval and data-use restrictions. MUSC and PRISMA indicators are not publicly
redistributable by the authors. Analysis code is available from the corresponding authors upon reasonable request,
subject to institutional approvals and data-use restrictions.


\bibliographystyle{elsarticle-harv}
\bibliography{references}

\end{document}